\pdfoutput=1
\documentclass[11pt]{article}

\usepackage[final]{acl}

\usepackage{times}
\usepackage{comment}
\usepackage{latexsym}
\usepackage{booktabs}
\usepackage{graphicx}
\usepackage{makecell}
\usepackage{hyperref}
\usepackage{multirow}
\usepackage[T1]{fontenc}
\usepackage[utf8]{inputenc}

\usepackage{microtype}

\usepackage{inconsolata}
\usepackage{graphicx}
\usepackage{subfig}
\usepackage{graphicx}

\title{Can Smaller LLMs do better? Unlocking Cross-Domain Potential through Parameter-Efficient Fine-Tuning for Text Summarization}

\author{Anum Afzal \\
  Technical University of Munich \\
  \texttt{anum.afzal@tum.de} \\ 
  \And
  Mehul Kumawat \\
Technical University of Munich \\
  \texttt{mehul.kumawat@tum.de} 
  \AND
  Florian Matthes \\
  Technical University of Munich \\
  \texttt{matthes@tum.de} \\ \\}

\begin{document}
\maketitle
\begin{abstract}
Large Language Models (LLMs), being generic task solvers, are versatile. However, despite the vast amount of data they are trained on, there are speculations about their adaptation capabilities to a new domain. Additionally, the simple fine-tuning of the model to incorporate knowledge of a new domain is computationally expensive and time-consuming. This becomes more challenging when the domain in question is also low-resource, and labeled data is unavailable. We leverage parameter-efficient fine-tuning techniques (PEFTs) on high-resource datasets to address these challenges to improve performance on unseen low-resource domains. Throughout our experiments, we evaluate whether intrinsic linguistic commonalities between datasets can be leveraged for efficient domain adaptation. We benchmark six PEFTs with \texttt{Llama-3-8B-Instruct} on 14 training datasets from the Scientific, Medical, Legal, and News domains for a Text Summarization task. Our experiments show that for low-resource domains, inference using Within-Domain Adapters can achieve better performance than Few-Shot as well as a much larger \texttt{Llama-3-70B-Instruct}. Lastly, in the absence of Within-Domain Adapters, we explore the concept of using Cross-Domain Adapters as well as the strategic combinations of adapters to leverage intrinsic language similarities across domains, facilitating better adaptability and performance in low-resource settings.
\end{abstract}

\section{Introduction}

Since the introduction of Large Language Models (LLMs), the field of Natural Language Processing (NLP) has advanced drastically. Given a large number of parameters, LLMs are very good at NLP tasks like commonsense reasoning \cite{yang2023commonsense}, multiple-choice question answering \cite{robinson2023mcqa}, and story generation \cite{cao2023storygeneration} in a zero-shot setting. However, these generic task solvers often lag when it comes to domain specialization. When an LLM is provided with a new domain, it does not adapt well to it and tends to hallucinate \cite{bang2023multitaskmultilingualmultimodalevaluation}. While several domain adaptation techniques can be used to overcome the domain-shift, this requires labeled training data. Unfortunately, these new domains are often also low-resource and lack labelled training data. Furthermore, full fine-tuning requires high compute, a large amount of labeled data, and may lead to catastrophic forgetting \cite{luo2025empiricalstudycatastrophicforgetting} in the model. Recent work has focused on adapting LLM to a new domain using PEFTs for tasks like Question Answering \cite{phogat2024finetuningsmallerlanguagemodels, chung2024efficientindomainquestionanswering}, Text Classification \cite{yang-etal-2022-parameter, zhou-etal-2022-making}, and reasoning tasks \cite{balne2024parameterefficientfinetuning}. However, the performance of PEFT on a Text Summarization task is still an open research question. 

\noindent We address this research gap by investigating whether parameter-efficient fine-tuning (PEFT) on a similar high-resource domain can help mitigate the effects of domain shift when working with a low-resource domain. Specifically, we explore PEFT techniques \cite{petl_4nlp, rebuffi2017learningmultiplevisualdomains} under two settings: \textbf{Within-Domain, where a high-resource dataset from the same domain is available, and Cross-Domain, where no such in-domain high-resource data exists.}

We opt for PEFT techniques as a domain adaptation technique instead of In-Context Learning (ICL) \cite{huang2022incontextlearningdistillationtransferring} because the articles to be summarized are often long and would require an extremely long context window. Contemporary LLM's support very long context window but require extensive hardware for operation. Additionally, it is also not clear how well the attention is distributed over an extremely long context window\cite{shi-penn-2025-semantic}.

\noindent In our methodology, fine-tuning using one PEFT on a given dataset would give an adapter\footnote{(IA)$^3$ is not considered to be an adapter because, unlike other techniques, it adds learned vectors to the attention mechanism. However, for the sake of simplicity, we consider it to be one.} that contains information related to a specific domain. We additionally focus on leveraging one or a combination Adapters trained on high-resource domains using PEFTs for text summarization, so LLMs generalize well on unseen low-resource domains. Given the modular nature of adapters, they can be added or removed for inference on a given domain with no additional overhead.

\noindent Given the large number of PEFTs to choose from, we benchmark six contemporary PEFTs including \texttt{AdaLoRA} \cite{zhang2023adaloraadaptivebudgetallocation}, \texttt{(IA)$^3$} \cite{liu2022ia3}, \texttt{LoHA} \cite{hyeonwoo2023loha}, \texttt{LoKr} \cite{yeh2024lokrlycoris}, \texttt{LoRA} \cite{hu2021loralowrankadaptationlarge}, and \texttt{OFT} \cite{qiu2024oft}. Our experiments span 14 datasets across four domains: Medical, Scientific, Legal, and News. We analyze the performance of these PEFTs both individually and in combination to assess their generalization ability under different settings. We hypothesize that although language data originate from diverse domains, these domains are intrinsically related, and this relationship can be leveraged to improve the generalization capability of large language models (LLMs) in low-resource scenarios, both Within-Domain (WID) and Cross-Domain (CD). Our main contributions are as follows:

 %We fine-tune the LLaMA3-B-Instruct model for the task of summarization on 14 datasets in the scientific domain - Arxiv \cite{arxiv_pubmed}, Elsevier \cite{elsevier_dataset}, and SciTLDR \cite{scitldr_dataset}; medical domain - CORD-19 \cite{Wang2020CORD19TC}, MSLR \cite{mslr_dataset}, PubMed \cite{arxiv_pubmed}, and SciLay \cite{scilay}; legal domain - BillSum \cite{kornilova-eidelman-2019-BillSum}, Eur-Lex-Sum \cite{aumiller-etal-2022-eur} and MultiLex-Sum \cite{Shen2022MultiLexSum}; and news domain - CNN/DM \cite{cnn_dm_dataset}, Multinews \cite{alex2019multinews}, Cornell Newsroom \cite{cornewsroom_dataset}, and Extreme Summarization (XSum) \cite{xsum_dataset}.  
\begin{itemize}

    \item We provide a Benchmark over six PEFTs and their domain-specific performance spanning over 14 datasets covering 4 domains.
    \item We demonstrate that a smaller LLM such as \texttt{Llama3-8B-Instruct}, can be successfully adapted to diverse domains using Adapters trained on similar high-resource domains, evaluating the usage of Within-Domain Adaptation.
    \item We provide a stance on using a combination of Adapters trained on high-resource datasets, which can improve performance on previously unseen low-resource domains.
    \item We explore the performance of Cross-Domain adapters when Within-Domain adapters may not be available for a low-resource dataset. Our experiments provide novel insights into cross-domain generalization, which is a relatively unexplored area in Domain Adaptation research.
   
\end{itemize}

\section{Related Work}
% @mehul. I would leave this to you. Just make sure you include the latest references, and maybe do a quick round of literature review to see if something new came up. I will review this section in the end.

\section{Domains}

 % Adjust row height for readability
        \begin{table}[ht]
            \centering
            \resizebox{\columnwidth}{!}{
                     \begin{tabular}{lrrrr}
                    %\hline
                     & \multicolumn{2}{c}{\textbf{Train}} & \multicolumn{2}{c}{\textbf{Test}} \\
                     \cmidrule(lr){2-3}  \cmidrule(lr){4-5}
                    
                      \textbf{Dataset} & \multicolumn{1}{c}{\# W} & \multicolumn{1}{c}{ \# Sum W} & \multicolumn{1}{c}{\# W} & \multicolumn{1}{c}{ \# Sum W} \\
                    \midrule
                    \multicolumn{5}{c}{Scientific} \\
                    \midrule
                    Arxiv                 & 1686.98 & 1140.18 & 2545.01 & 172.31  \\
                    Elsevier              & 1427.81 & 206.57 & 1835.53 & 215.81       \\ 
                    SciTLDR               & 153.41 & 23.03 & 137.10 & 79.68        \\
                    \midrule
                    
                    \multicolumn{5}{c}{Medical}\\
                    \midrule
                    CORD-19               &  2275.53 & 246.95  & 2967.36 & 272.90        \\ 
                    MSLR (MS\^{}2)        & 1453.44 & 66.77 & 7741.43 & 66.60        \\ 
                    PubMed                & 905.92 & 165.59  & 1127.95 & 173.33      \\ 
                    SciLay                &  4239.51 & 161.32 & 4712.58 & 164.00       \\ 
        
                    \midrule
                    \multicolumn{5}{c}{Legal}\\
                    \midrule
                    BillSum               &  1073.30 & 159.34  & 1123.43 & 176.17        \\ 
                    Eur-Lex-Sum           & 7478.22 & 1053.74  & 13788.87 & 1318.67          \\ 
                    Multi-LexSum          & 10910.32 & 309.16  & 7464.81 & 253.59 \\ 
        
                    \midrule
                    \multicolumn{5}{c}{News}\\
                    \midrule
                    CNN/DM                & 306.35 & 52.53 &  348.64 & 59.86       \\ 
                    Multi-news            & 448.26 & 215.47 & 501.71 & 226.20       \\ 
                    Newsroom              & 131.42 & 29.24 &  166.57 & 28.65         \\ 
                    XSum                  & 86.08 & 23.13 & 110.20 & 24.43        \\ 
                    \bottomrule
                \end{tabular}

            }
            \caption{The average tokens in articles and summary of train and test split across datasets, calculated using Llama3-8B tokenizer. \# W refers to average number of tokens in articles, and \# Sum W refers to the average number of tokens in reference summaries.}
            \label{tab:data_tokens}
        \end{table}

To improve the generalization abilities of LLM, we evaluate our methodology on four domains: scientific, medical, legal, and news. We utilize 14 datasets for training adapters across these domains, and 4 holdout validation sets consisting of real-life datasets from each domain.
\subsection{Training Datasets}    
    \paragraph{Scientific:}
        In the scientific domain, we train adapter layers with three datasets: (i) \textbf{ArXiv} summarization dataset  \cite{arxiv_pubmed} - long, structured articles from Arxiv journals; (ii) \textbf{Elsevier} dataset \cite{elsevier_dataset} - collection of CC-BY licensed scientific papers; and (iii) \textbf{SciTLDR} \cite{scitldr_dataset} - dataset with ultrashort summaries of AICs\footnote{AIC: Abstract, Introduction, and Conclusion} of scientific articles.  
    \paragraph{Medical:}
        For domain adaptation in the medical domain, we leveraged four datasets: (i) \textbf{CORD-19} \cite{Wang2020CORD19TC} - COVID-19 medical research articles with summaries; (ii) \textbf{MSLR} \cite{mslr_dataset} - medical evidence summarized as literature reviews; (iii) \textbf{PubMed} \cite{arxiv_pubmed} -  long, structured articles from PubMed scientific journals; and (iv) \textbf{SciLay} \cite{scilay} - articles from biomedical domain, which aim to provide simplified summaries for complex articles. 
    \paragraph{Legal:}
        We train Adapters to summarize legal and constitutional documents from (i) \textbf{BillSum} dataset \cite{kornilova-eidelman-2019-BillSum} - summaries of US Congressional and California State bills; (ii) \textbf{Eur-Lex-Sum} dataset \cite{aumiller-etal-2022-eur} - consists of human-authored summaries of European Union legal acts - we work with the English texts only; and (iii) \textbf{Multi-Lex} dataset \cite{Shen2022MultiLexSum} - a collection of multiple legal documents for summarization. 
    \paragraph{News:}
        In the news domain, we included four datasets - (i) \textbf{CNN/DM} dataset \cite{cnn_dm_dataset} - English language corpus authored by CNN and DM journalists; (ii) \textbf{Multinews} dataset \cite{alex2019multinews} - expert summaries of articles from \textit{\url{newser.com}}  (iii) \textbf{Cornell Newsroom} \cite{cornewsroom_dataset} - collection of news articles with summaries by professionals for advanced summarization; and (iv) \textbf{XSum} dataset \cite{xsum_dataset} - scrapped online articles from BBC news from abstractive summarization.
    
    \noindent \autoref{tab:data_tokens} shows the average tokens from each dataset utilized in training.
        
    \subsection{Holdout Validation Datasets}

        We compile one holdout validation dataset of 100 samples for each domain, which we show in \autoref{tab:hvd_data_tokens}. To avoid contamination, we make sure that the articles in our validation set are dated to be after \texttt{Llama 3} was released. This sanity check ensures that these articles were not part of the model's training regime, and thus these datasets can be considered low-resource. Details of our validation datasets are as follows:
 
 \paragraph{Scientific:} We evaluate ACLSum dataset \cite{takeshita2024aclsumnewdatasetaspectbased}, which covers multiple aspects of summarization, including challenges, approach, and outcome of scientific articles.
\paragraph{Medical:} We scraped data from BMJ Medical Research\footnote{\url{https://www.bmj.com/research/research}} to compile articles from last year, with the abstract as summaries.
\paragraph{Legal:} We use the legal cases from India and the UK with abstraction summaries compiled by \citealt{unseen_legaldata}
 \paragraph{News:} We fetch the summaries of the current affairs from the last week of 2024, where the summaries are expert written, collected from InShorts\footnote{\url{ https://inshorts.com/en/read}}.

            \begin{table}[ht]
            \centering
            
                \begin{tabular}{lrr}
                    
                    \textbf{Domain} &  \multicolumn{2}{c}{\textbf{Avg. Tokens }} \\ 
                    \midrule
                    & {Article} & {Summary} \\
                    \hline
                      Scientific & 1131.38 & 80.08         \\  %\cline{2-5}
                    Medical &   6059.04 & 587.40      \\ %\cline{2-5}
                    Legal  & 4281.12 & 906.04         \\ %\cline{2-5}
                    News & 509.29 & 74.72      \\
                    \hline
                \end{tabular}
            
            \caption{The average tokens in article and summary for our Holdout Validation Datasets using Llama3-8B tokenizer.}
            \label{tab:hvd_data_tokens}
        \end{table}

\section{Methodology}
\begin{figure*}[ht]
    \centering
    \subfloat[ Pipeline for Selecting a Domain Adapter. For each dataset, we train all PEFTs, and the best-performing PEFT is selected as the dataset representative. A similar approach is used to define a Domain representative which could be a single dataset adapter or a combination of datasets.]{{\includegraphics[width=10cm]{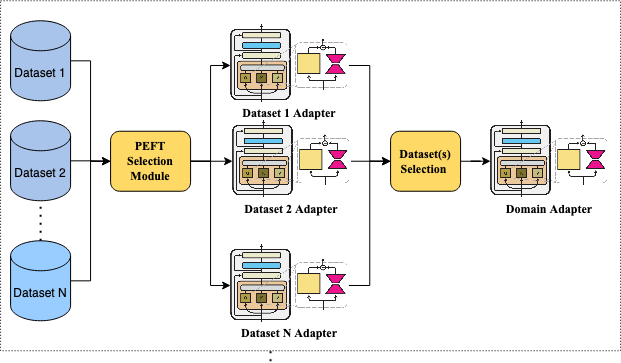}  \label{fig:method-a}}} 
    \qquad
    \subfloat[Using the Domain Adapters from (a), we test out Domain Adapters on the Holdout Validation dataset. This image illustrates the example of the evaluation Scientific Holdout Dataset in within and cross-domain settings.]{{\includegraphics[width=5cm]{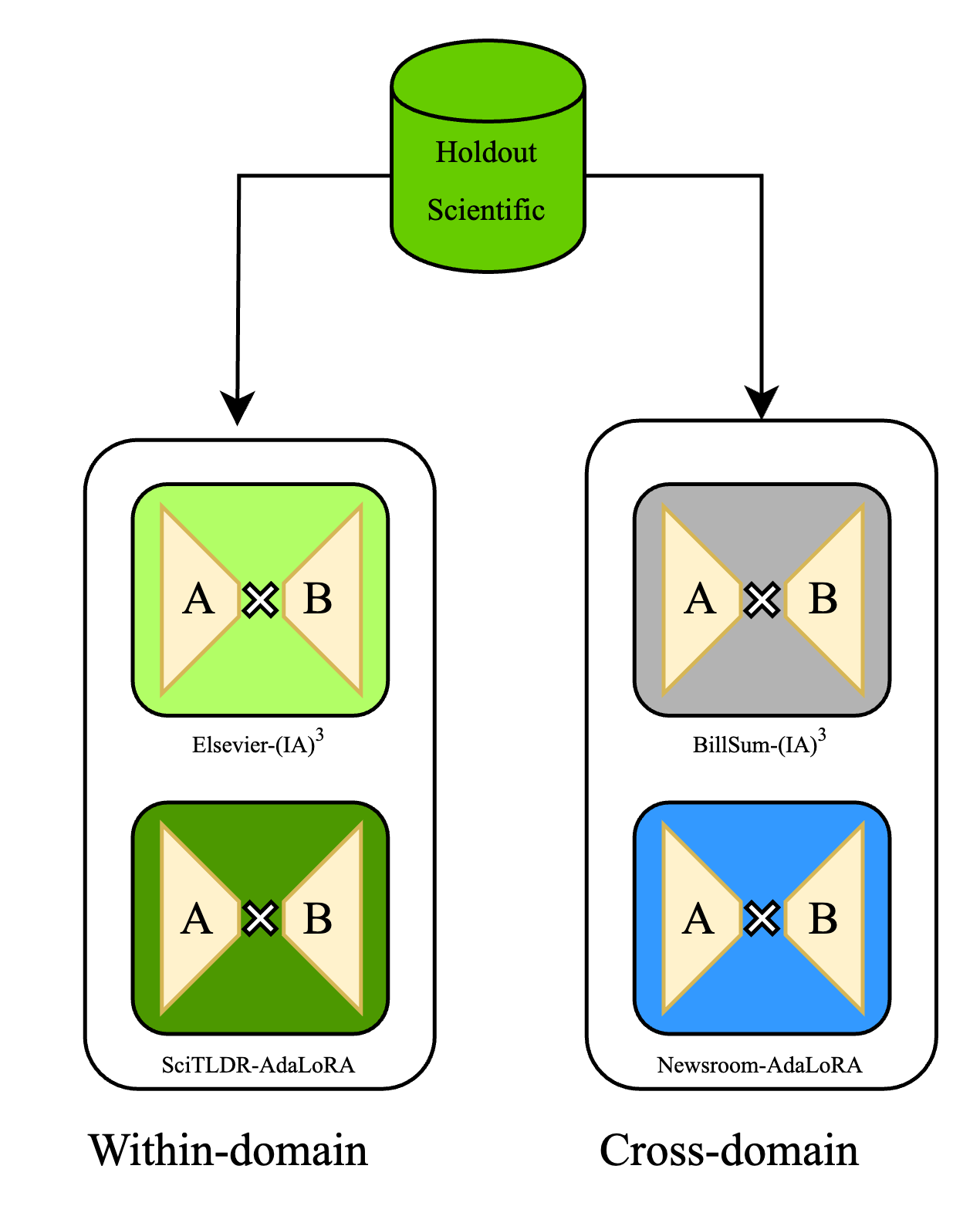} \label{fig:method-b} }}
    \caption{Our methodology focuses on 1) benchmarking state-of-the-art adapters on all datasets from a domain, to select adapters that can be used as domain representatives. and 2) Using the top-3 adapters from each domain on our Holdout Validation datasets under Within-Domain and Cross-Domain Settings. }%
    \label{fig:methodology}%
\end{figure*}
Given the unavailability of labeled training data in low-resource domains, we hypothesize that Adapters trained on higher-resource domains could help LLM generalize better. Given the modular nature of Adapter, they can be added to an LLM under the following scenarios, which we also illustrate in Figure \autoref{fig:method-b}:

\paragraph{Within-Domain:} We use one or a combination of adapters trained on the datasets from the same domain and test them on their respective holdout validation dataset. For example, we test the scientific holdout validation dataset on one or more Adapters trained on our high-resource scientific datasets. 

\paragraph{Cross-Domain:} We use one or a combination of Adapters trained on different domains to improve the LLM performance performance on our holdout validation dataset that we consider to be low-resource. For instance, we use one or a combination of Adapters from other domains.

\subsection{Dataset Adapters}
 
Given the large number of PEFT methods available, we conduct preliminary experiments to benchmark the performance of several state-of-the-art techniques, including \texttt{AdaLoRA} \cite{zhang2023adaloraadaptivebudgetallocation}, \texttt{(IA)$^3$} \cite{liu2022ia3}, \texttt{LoHA} \cite{hyeonwoo2023loha}, \texttt{LoKr} \cite{yeh2024lokrlycoris}, \texttt{LoRA} \cite{hu2021loralowrankadaptationlarge}, and \texttt{OFT} \cite{qiu2024oft}. As shown in Figure~\autoref{fig:method-b}, we evaluate each PEFT method across all datasets. These initial experiments offer insight into how PEFT performance varies depending on the dataset and which PEFT performs best for a given dataset.

\subsection{Domain Adapters}

We select the top-3 dataset-specific adapters from each domain as domain representatives, which are subsequently used in Within-Domain and Cross-Domain experiments on the Holdout Datasets. To evaluate domain generalization, we test all possible combinations of one or more adapters within the same domain. These combinations are constructed by merging adapters trained on datasets from the same domain, ensuring that none have been exposed to the specific test dataset—for example, evaluating the combination of \textit{Elsevier-AdaLoRA} and \textit{SciTLDR-AdaLoRA} on Arxiv’s test set. This process allows us to identify adapter configurations that effectively represent the broader domain in cross-domain scenarios.

%\subsection{In-Domain vs Cross-Domain}
%Our analysis of the holdout validation dataset shows how the in-domain and cross-domain settings perform for a low-resource dataset. Similar to our previous experiments,  we try combinations of one or more Adapters, which we illustrate in \autoref{tab:val-experiment-combination}. For example, the holdout validation dataset of the Scientific domain is evaluated by the combination of PubMed-LoRA and BillSum-LoKr from medical and legal domains, respectively, to represent a multi-adapter setting.

%\begin{table}[h]

%  \centering
%  \resizebox{\columnwidth}{!}{
%  \begin{tabular}{ccccc}
%    \toprule
%      Test Domain & Adapters & Type & DA & \\
%    \toprule
%    Scientific &{SciTLDR-AdaLoRA}  & single& {WID}\\
%    \midrule
%    \multirow{2}{*}{Medical}  & {BillSum-LoKr}  & \multirow{2}{*}{multi}&  \multirow{2}{*}{CD} \\
%     & CNNDM-AdaLoRA  & &  \\
%    \midrule
% Legal &{MultiLex-LoRA}& single & {WID} \\
% \midrule
 % News &{SciTLDR-AdaLoRA} &single& {CD}\\
% News &{CNNDM-AdaLoRA} &single& {WD}\\
% \midrule
%\multirow{3}{*}{Scientifc}  & {PubMed-LoRA}  & \multirow{3}{*}{multi}&  \multirow{3}{*}{CD} \\
%     & {BillSum-LoKr}  & &  \\
%     & {CNNDM-AdaLoRA}  & &  \\
%    \midrule
 % Legal & {} & single & WID\\
 %Legal & {PubMed-LoRA} & multi & CD\\
% & {Arxiv-LoKr}  & &  \\
 
 %   \bottomrule
 % \end{tabular}
 % }
 
 %  \caption{An illustration of a few combinations of within-domain and cross-domain combinations used in our experiments with Holdout Validation set where WID = Within-Domain and CD = Cross-Domain.}
 %   \label{tab:val-experiment-combination}
 %\end{table}

 \begin{table}[h]
    \centering
    % \scalebox{0.92}
    %\resizebox{\columnwidth}{}
    {    
            \begin{tabular}{llc}
                \hline
                \textbf{Domain} & \textbf{Dataset} & \textbf{Best PEFT}\\
                
                \hline
                % SCIENTIFIC ROWS
                \multirow{3}{*}{Scientific} 
                & Arxiv         & \textbf{LoKr*} \\ %\cline{2-8} 
                & Elsevier      & AdaLora* \\ %\cline{2-8} 
                & SciTLDR       & AdaLora* \\ \hline 
                
                % MEDICAL ROWS
                \multirow{4}{*}{Medical} 
                & CORD19        & AdaLora* \\ %\cline{2-8}
                & MSLR          & {(IA)$^3$} \\ %\cline{2-8}
                & PubMed        & \textbf{LoRA*} \\ %\cline{2-8}
                & Sci Lay       & {OFT*}  \\ \hline
                
                % LEGAL ROWS
                \multirow{3}{*}{Legal} 
                & BillSum       & \textbf{LoKr*} \\ %\cline{2-8}
                & Eur-Lex-Sum   & {OFT*} \\ %\cline{2-8} 
                & Multi-Lex     & {LoRA*}  \\ \hline

                % NEWS ROWS
                \multirow{4}{*}{News} 
                & CNN/DM            & \textbf{AdaLora*} \\ %\cline{2-8} 
                & Multi News        & {OFT*} \\ %\cline{2-8} 
                & Newsroom          & AdaLora* \\ %\cline{2-8}
                & XSum              & {LoKr} \\ \hline 
                
            \end{tabular}
        }
    %\vspace{3mm}
    \caption{Best performing PEFTs for individual datasets. * implies that the combination is among top-3 adapters within a given domain. Our single dataset adapters outperform the combination of one or more dataset adapters. See Appendix \ref{app:domain-adapters} for a detailed section on individual PEFT scores and ranks. }
    \label{tab:pefts_rank}
\end{table}
     
\subsection{Evaluation Metrics:}
\label{meth:eval-metrics}
For evaluation, we rely on more than one metric for a broader coverage, including ROUGE \cite{lin-2004-rouge}, BERTScore \cite{zhang2020bertscoreevaluatingtextgeneration}, BLEU \cite{bleuuu}, METEOR \cite{banerjee-lavie-2005-meteor}, and FActScore \cite{min2023factscorefinegrainedatomicevaluation} for the evaluation of PEFTs and Datasets. A large number of metrics makes ranking challenging so we use the Borda Count Method%\footnote{\url{https://en.wikipedia.org/wiki/Borda_count}}
\cite{bordacount} to get the ranks for PEFTs within a dataset, and datasets within a domain. For our Holdout datasets, we also do annotations using domain experts.

            \begin{table*}[ht]
            \centering
             \resizebox{\textwidth}{!} 
            {
             
                \begin{tabular}{llrrrrrr}
                    \toprule 
                     \textbf{{Setting}} & \textbf{Combined PEFTs} & \multicolumn{1}{c}{\textbf{ROUGE}} & \textbf{BERTScore} & \textbf{BLEU} & \textbf{METEOR} & \textbf{FActScore}\\
%-----------------------------------------------SCIENTIFIC--------------------------------------------------------
                        \toprule
                        \multicolumn{7}{c}{\textit{{Scientific Domain}}} \\
                        \toprule
                    % & Zero-Shot   & \textit{0.2265} & \textit{0.8779} & \textit{0.0701} & \textit{\textbf{0.3974}} & \textit{0.9734} \\ %& \textit{28.3000} \\
                    % % \textit{0.3502} &	\textit{0.1397} &	\textit{0.2048} &	\textit{0.8766} &	\textit{0.0652} \\
                    % \hline
                    % \multirow{1}{*}{Within-Domain}
                    \multirow{3}{*}{WID}
                    % &\makecell[l]{Adapter 1 \\ + Adapter 2} 
                    % & 0.2445 & 0.0768 & 0.1513 & 0.8488 & 0.0375 \\ 
                    &\textbf{SciTLDR-AdaLoRA} & \textbf{0.2746} & \textbf{0.8849} & \textbf{0.1107} & 0.3913 & 0.9489\\ 
                    %\cline{2-7} %& 16.8300 \\ 
                    &Arxiv-AdaLoRA & 0.2481 & 0.8773 & 0.0829 & \textbf{0.4026} & 0.9503\\ 
                    %\cline{2-7} %& 16.8300 \\ 
                    & {Arxiv-LoKr + SciTLDR-AdaLoRA} & 0.1698 & 0.8467 & 0.0457 & 0.3467 & {0.8948} \\%& - \\

                    % \hline
                    \toprule
                    % Across Domain
                    \multirow{3}{*}{CD}
                    &{PubMed-LoRA +BillSum-LoKr} & 0.1967 & 0.8593 & 0.0568 & 0.3735 &  \textbf{0.9754} \\ 
                    %\cline{2-7}%& 35.1200 \\ 
                    &{PubMed-LoRA + CNNDM-AdaLoRA} & 0.1929 & 0.8570 & 0.0564 & 0.3635 & 0.9741 \\ 
                    %\cline{2-7}%& 35.3300 \\ 
                    &\makecell[l]{PubMed-LoRA + BillSum-LoKr +\\CNNDM-AdaLoRA} & 0.1814 & 0.8536 & 0.0512 & 0.3477 & 0.9501 \\ %& 35.0800 \\ 
                    
%-----------------------------------------------MEDICAL----------------------------------------------------------
                    \toprule
                        \multicolumn{7}{c}{\textit{{Medical Domain}}} \\
                        \toprule
                    % & Zero-Shot   & \textit{0.2575} & \textit{0.8488} & \textit{0.0588} & \textit{0.2095} & {\textbf{0.9586}} \\%& { 26.8125} \\
                    % \hline
      
                    % \multirow{1}{*}{Within-Domain}
                    \multirow{3}{*}{WID}
                    &\textbf{CORD19-OFT} & \textbf{0.2856} & \textbf{0.8606} & \textbf{0.0864} & \textbf{0.2417} & 0.9417 \\%& {34.1354} \\
                    %\cline{2-7}
                    &{PubMed-LoRA} & 0.2750 & 0.8485 & 0.0779 & 0.2186 & \textbf{0.9565} \\%& {} \\ 
                    %\cline{2-7}
                    &{CORD19-AdaLoRA + MSLR-IA3} & 0.2759 & 0.8590 & 0.0777 & 0.2315 & {0.9364} \\%& {} \\ 

                    % &\makecell[l]{SciTLDR-AdaLoRA \\ + Arxiv-AdaLoRA} 
                    % & 0.2445 & 0.0768 & 0.1513 & 0.8488 & 0.0375 \\ 
                    % divider
                    % \hline
                    \toprule
                    % Across Domain
                    \multirow{3}{*}{CD}
                    &{BillSum-LoKr + CNNDM-AdaLoRA} & 0.2421 & 0.8339 & 0.0704 & 0.2156 & {0.9191} \\%& {29.9565}\\ 
                    %\cline{2-7}
                    &{Arxiv-LoKr + BillSum-LoKr} & 0.2268 & 0.8250 & 0.0644 & 0.2029 & {0.8468} \\%& {28.7188}\\ 
                    %\cline{2-7}
                    &{Arxiv-LoKr + CNNDM-AdaLoRA} & 0.2159 & 0.8208 & 0.0605 & 0.1943 & {0.8108} \\%& {29.9565}\\ 

%-----------------------------------------------LEGAL------------------------------------------------------------
                    \toprule
                        \multicolumn{7}{c}{\textit{{Law Domain}}} \\
                        \toprule
                    % & Zero-Shot   & \textit{0.2360} & \textit{0.8522} & \textit{0.0210} & \textit{0.1648} & \textit{\textbf{0.9255}}  \\%& \textit{26.56} \\
                    % \hline

                    % \multirow{1}{*}{Within-Domain}
                    \multirow{3}{*}{WID}
                    &\textbf{MultiLex-LoKr} & \textbf{0.2411} & \textbf{0.8526} & \textbf{0.0216} & \textbf{0.1679} & 0.9166 \\%& 29.07 \\ 
                    %\cline{2-7}
                    &{BillSum-AdaLoRA} & 0.2381 & 0.8516 & 0.0190 & 0.1625 & \textbf{0.9171}\\
                    %\cline{2-7}
                    &{MultiLex-LoRA + BillSum-LoKr} & 0.2077 & 0.8401 & 0.0182 & 0.1465 & 0.9048\\ %& 32.7980 \\ 
            
                    % &\makecell[l]{SciTLDR-AdaLoRA \\ + Arxiv-AdaLoRA} 
                    % & 0.2445 & 0.0768 & 0.1513 & 0.8488 & 0.0375 \\ 

                    % divider
                    % \hline
                    \toprule
                    % Across Domain
                    \multirow{3}{*}{CD}
                    &{PubMed-LoRA + CNNDM-AdaLoRA} & 0.2146 & 0.8346 & 0.0191 & 0.1486 & 0.8893 \\%& 24.5217 \\ 
                    %\cline{2-7}
                    &{PubMed-LoRA + Arxiv-LoKr} & 0.2051 & 0.8190 & 0.0150 & 0.1445 & 0.6723 \\%& 13.5185 \\ 
                    %\cline{2-7}
                    &{Arxiv-LoKr + CNNDM-AdaLoRA} & 0.1900 & 0.8215 & 0.0161 & 0.1372 & 0.7895 \\%& 27.3895 \\ 
                    %\cline{2-7}
                    \toprule
%-----------------------------------------------NEWS-------------------------------------------------------------
                        \multicolumn{7}{c}{\textit{{News Domain}}} \\
                        \toprule
                    % & Zero-Shot   & \textit{0.2488} & \textit{0.8889} & \textit{0.0912} & \textit{0.4040} & \textit{\textbf{0.9768}} \\%& \textit{24.4000} \\
                    % \hline
      
                    % \multirow{1}{*}{Within-Domain}
                    \multirow{2}{*}{WID}
                    &\textbf{Newsroom-LoKr} & \textbf{0.2976} & \textbf{0.8868} & \textbf{0.1283} & \textbf{0.4438} & 0.8909 \\%& 28.5882 \\ 
                    %\cline{2-7}
                    &{Multinews-AdaLoRA} & 0.2604 & 0.8864 & 0.0976 & 0.4167 & \textbf{0.9700} \\
                    %\cline{2-7}
                    &{MultiNews-OFT + Newsroom-AdaLoRA} & 0.1866 & 0.8612 & 0.0665 & 0.3595 & {0.8306} \\ % & - \\ 
            
                    % &\makecell[l]{SciTLDR-AdaLoRA \\ + Arxiv-AdaLoRA} 
                    % & 0.2445 & 0.0768 & 0.1513 & 0.8488 & 0.0375 \\ 
      
                    % divider
                    % \hline
                    \toprule
                    % Across Domain
                    \multirow{3}{*}{CD}
                    &{PubMed-LoRA + BillSum-LoKr} & 0.2443 & 0.8747 & 0.1001 & 0.4260 & 0.9560 \\%& 36.9412 \\ 
                    %\cline{2-7}
                    &{Arxiv-LoKr + BillSum-LoKr} & 0.2348 & 0.8712 & 0.0955 & 0.4147 & 0.9447 \\%& 37.3647 \\ 
                    %\cline{2-7}
                    &{PubMed-LoRA + Arxiv-LoKr} & 0.2123 & 0.8533 & 0.0667 & 0.3764 & 0.9555 \\%& 36.1176 \\ 

                    % &\textbf{\makecell[l]{BillSum-(IA)$^3$ \\ + Newsroom-AdaLoRA} }
                    % & \textbf{0.3978} & \textbf{0.1667} & \textbf{0.2456} & \textbf{0.8806} & \textbf{0.1004} \\
                    % \cline{2-7}
                    
                    % &\makecell[l]{CORD19-OFT \\ + BillSum-(IA)$^3$} 
                    % & 0.3062 & 0.1250 & 0.1717 & 0.8621 & 0.0549 \\ %\hline
                    % \cline{2-7}
                    % &\makecell[l]{CORD19-OFT \\ + BillSum-(IA)$^3$ \\ + Newsroom-AdaLoRA} 
                    % & 0.3011 & 0.1220 & 0.1697 & 0.8602 & 0.0531  \\ 
                    \toprule
                \end{tabular}
            }
            %\vspace{3mm}
            \caption{Performance of one or more domain-specific adapters in Within domain and Cross Domain settings inferred on the Holdout validation dataset for all domains. Note: ROUGE is the geometric mean of R1, R2, and RL. WID and CD refer to Within-Domain and Cross-Domain, respectively.}
            \label{tab:within_cross_scientific_unseen_test_data}
        \end{table*}

\subsection{Baseline}

We use \texttt{Meta-Llama-3-8B-Instruct} in zero-shot and ICL (In-Context Learning) settings, and \texttt{Meta-Llama-3-70B-Instruct} in zero-shot as our baseline. We generated summaries with the prompts shown in \autoref{subsec:prompts} for all holdout validation datasets and compared them with our within-domain and cross-domain settings.

\subsection{Experimental Settings}
Recent studies show that in a zero-shot setting, smaller LLMs suffer more from the domain shift as compared to their bigger counterparts \cite{afzal-etal-2024-adapteval}. Based on this research gap, we used \texttt{Llama3-8B-Instruct} model as an experimental choice for our Methodology. We fine-tune the \texttt{Llama3-8B-Instruct}\footnote{We chose the Llama3-8B-Instruct to avoid potential knowledge contamination, as it had not been exposed to our holdout validation datasets. Newer models like Qwen 2.5, GPT-4, or DeepSeek could risk including prior exposure to our validation datasets during their training phase. We aim to maintain the integrity and originality of our validation process.} using six PEFTs - AdaLoRA, (IA)$^3$, LoHA, LoKr, LoRA, and OFT, available in the \textit{peft} library from the HuggingFace on all datasets. We perform \textit{Instruction-tuning} through \textit{Adapter Training}, attaching one of the above-mentioned PEFTs to the self-attention blocks of the Llama3, allowing for efficient training of less than 1\% of the model's parameters while keeping the original model frozen. For each domain, we used a domain-specific prompt (see Appendix \ref{subsec:prompts}) alongside content and summaries using 1000 samples \cite{10.5555/3666122.3668522}. Each fine-tuning instance results in Adapter layers, which we store and use as individual modules. During generation, we use a maximum of 256 new tokens. See \autoref{app:tech-details} for details of fine-tuning and generation hyperparameters.

\section{Results and Discussion}
\label{sec:results}
\subsection{PEFT Benchmark over Datasets}

We evaluate the effectiveness of each PEFT by testing it on the datasets they were originally trained on. To ensure a fair comparison, we rank the adapters using the Borda Count method based on multiple evaluation metrics, as described in \autoref{meth:eval-metrics}. The results, summarized in \autoref{tab:pefts_rank}, reveal that PEFT performance is highly dataset-dependent, highlighting the need for dataset-specific evaluations rather than relying solely on average scores. Interestingly, while no single PEFT consistently dominates across all datasets, \texttt{AdaLoRA} demonstrates superior performance on a majority of them, particularly within the Scientific and News domains. For instance, it outperforms other methods on CORD19, SciTLDR, Elsevier, CNN/DM, and Newsroom datasets. This suggests that \texttt{AdaLoRA} is particularly well-suited for abstractive summarization tasks in data-rich or heterogeneous domains. Other methods, such as \texttt{LoKr} and \texttt{OFT}, perform strongly on specific datasets like Arxiv, BillSum, and Multi-News, indicating that different PEFTs may capture domain-specific structures differently. See Appendix \ref{app:peft-performance} for a detailed benchmark.

%These findings support the idea of selecting top-performing adapters on a per-dataset basis. Doing so allows us to construct domain-representative adapter sets for broader Within-Domain and Cross-Domain evaluations. Overall, the variability in PEFT effectiveness underscores the importance of fine-grained benchmarking in domain-adaptive NLP settings. 

\noindent Our findings also reveal that, across all domains, the top-3 ranked adapters are individual (single) adapters rather than combinations. This suggests that, contrary to expectations, merging multiple dataset-specific adapters within the same domain does not lead to significant performance improvements. In fact, individual adapters often capture domain-specific patterns more effectively, likely due to reduced interference and greater specialization. These observations indicate that, for Within-Domain applications, a well-chosen single adapter may suffice.

%negatively affects the LLM performance on the downstream task.
        \begin{table}[th]
            \centering
            
             \resizebox{\columnwidth}{!} % 0.8\linewidth => we finish in 8 pages. 
            {
            \begin{tabular}{clcccc}
                \hline
                
                \textbf{{Setting}} &
                \textbf{\makecell[l]{Training \\Dataset}} & \textbf{\makecell[c]{Vocab\\ Overlap}} & \textbf{\makecell[c]{TF-IDF\\Overlap}} & \textbf{KL Div} & \textbf{\makecell[c]{Contextual\\Overlap}} \\
                \toprule
                \multicolumn{6}{c}{Scientific} \\
                \toprule
                \multirow{3}{*}{WID} 
                 &Arxiv     & 42.22 & 43.96 & 12.30 & 0.26 \\
                           & Elsevier  & 40.45	&41.64	&12.97	&0.23 \\
                           & SciTLDR  & 16.75 & 18.31 & 4.95  & 0.35 \\
                
                \toprule
                 \multirow{3}{*}{CD}  
                 &BillSum   & 26.25 & 28.64 & 12.81 & 0.16 \\
                           & CNN/DM    & 18.14 & 19.65 & 16.56 & 0.08 \\
                           & PubMed    & 34.29 & 37.01 & 14.14 & 0.20 \\
                \toprule
                \multicolumn{6}{c}{Medical} \\
                \toprule
                \multirow{4}{*}{WID}
                 &CORD19 & 31.58 & 48.69 & 6.18 & 0.37 \\ %\cline{2-6}
                          & MSLR & 39.84 & 50.83 & 5.26 & 0.40 \\ %\cline{2-6}
                          & SciLay & 34.66	&46.80	&7.65	&0.35 \\ %\cline{2-6}
                          & PubMed & 22.42	&40.11	&7.08	&0.28 \\ %\cline{2-6}
                \toprule
                \multirow{3}{*}{CD}
                 &Arxiv & 20.90 & 34.64 & 10.50 & 0.21 \\ %\cline{2-6}
                          & BillSum & 14.71 & 26.88 & 8.78 & 0.17 \\ %\cline{2-6}
                          & CNN/DM & 8.89 & 16.12 & 13.62 & 0.11 \\ 
                
                \toprule
                \multicolumn{6}{c}{Legal} \\
                \toprule
                \multirow{3}{*}{WID}
                & BillSum & 23.15 & 29.55 & 5.85 & 0.26 \\ %\cline{2-6}
                           & EurLex & 42.99	&47.58	&4.26	&0.28 \\ %\cline{2-6}
                           & MultiLex & 32.71 & 33.61 & 8.11 & 0.39 \\ %\cline{2-6}
                \toprule
                \multirow{3}{*}{CD}
                 &Arxiv & 27.67 & 31.28 & 12.04 & 0.17 \\ %\cline{2-6}
                          & CNN/DM & 15.36 & 19.03 & 11.45 & 0.14 \\ %\cline{2-6}
                          & PubMed & 23.95 & 29.23 & 11.75 & 0.19 \\ 
                
                \toprule
                \multicolumn{6}{c}{News} \\
                \toprule
                \multirow{4}{*}{WID}
                 &CNN/DM & 27.78	&28.87	&11.21	&0.12 \\ %\cline{2-6}
                          &Multinews & 37.38 & 39.13 & 10.34 & 0.13 \\ %\cline{2-6}
                          & Newsroom & 23.78	&25.04	&10.87	&0.12 \\ %\cline{2-6}
                          & XSum & 16.49	&16.89	&10.61	&0.11 \\ %\cline{2-6}
                \toprule
                \multirow{3}{*}{CD}
                 &Arxiv & 35.49 & 35.78 & 15.20 & 0.11 \\ %\cline{2-6}
                           & BillSum & 31.37 & 32.98 & 10.64 & 0.11 \\ %\cline{2-6}
                           & PubMed & 36.49 & 38.28 & 13.42 & 0.12 \\
                \hline
            \end{tabular}
            }
            \caption{Linguistic and semantic similarity metrics between holdout validation datasets and training datasets. KL div refers to Kullback–Leibler divergence.}
            
            \label{tab:domain_similarity}
        \end{table}

\subsection{Cross-Domain vs Within-Domain}

We evaluated and summarized the selected domain adapters in Within-Domain (WID) and Cross-Domain (CD) settings on four holdout validation datasets in \autoref{tab:within_cross_scientific_unseen_test_data}. Overall, within-domain combinations of adapters yield performance that is largely comparable to, and in some cases surpasses, that of cross-domain adapters. Notably, within-domain configurations often produce higher FActScores, indicating a stronger ability to preserve factual consistency in generated summaries. Performance across domains also highlights the nuanced behavior of adapter combinations. While certain cross-domain combinations exhibit synergy, others suffer from diminished performance, potentially due to conflicting adaptation signals or overfitting to non-representative data. Additionally, adapter configurations involving more than two components generally show diminishing returns or even degradation in scores, emphasizing the importance of carefully curating cross-domain blends.

%These findings reinforce the utility of cross-domain transfer for general-purpose summarization tasks, while also underlining the continued relevance of domain specialization in fields like biomedicine. They also demonstrate the potential for cross-domain adapters to enhance factual correctness, a key concern in high-stakes summarization tasks.

            \begin{table*}[ht]
            % \begin{table*}[htbp]
            \centering
            
             \resizebox{0.85\textwidth}{!} 
            {
                % \begin{tabular}{llrrrrrr}
                \begin{tabular}{llrrrrrr}
                    \hline 
                     \textbf{Model} & \textbf{{$k$}} & \multicolumn{1}{c}{\textbf{ROUGE}} & \textbf{BERTScore} & \textbf{BLEU} & \textbf{METEOR} & \textbf{FActScore}\\
%-----------------------------------------------SCIENTIFIC--------------------------------------------------------
                    \hline
                    \multicolumn{7}{c}{\textit{{Scientific Domain}}} \\
                    \hline

                    \multirow{2}{*}{{Llama3-8B-Instruct}}
                    % &\makecell[l]{Adapter 1 \\ + Adapter 2} 
                    % & 0.2445 & 0.0768 & 0.1513 & 0.8488 & 0.0375 \\ 
                    &{0} & 0.2265 & {0.8779} & {0.0701} & 0.3974 & \textbf{0.9735}\\ 
                    % &1 & 0.2035 & 0.8692 & 0.0582 & 0.3829 & 0.9062\\ 
                    &2 & 0.1214 & 0.8353 & 0.0504 & 0.1700 & 0.5131\\ 
                    \midrule
                    \multirow{1}{*}{{Llama3-70B-Instruct}}
                    &{0} & 0.2311	&0.8766	&0.0696	&\textbf{0.4108} &  0.9025 \\ 
                    \midrule 
                    {SciTLDR-AdaLoRA} & - &\textbf{0.2746} & \textbf{0.8849} & \textbf{0.1107} & 0.3913 & 0.9489\\ 
                    
%-----------------------------------------------MEDICAL----------------------------------------------------------
                    \hline
                    \multicolumn{7}{c}{\textit{{Medical Domain}}} \\
                    \hline
                    \multirow{2}{*}{{Llama3-8B-Instruct}}
                    % &\makecell[l]{Adapter 1 \\ + Adapter 2} 
                    % & 0.2445 & 0.0768 & 0.1513 & 0.8488 & 0.0375 \\ 
                    &{0} & {0.2575} & {0.8488} & {0.0588} & {0.2095} & {0.9586}\\ 
                    % &1 & 0.2565 & 0.8469 & 0.0526 & 0.2034 & 0.9601\\ 
                    &2 & 0.2572 & 0.8486 & 0.0527 & 0.2055 & \textbf{0.9661}\\ 
                    \midrule
                    % \hline
                    \multirow{1}{*}{{Llama3-70B-Instruct}}
                    &{0} & 0.2527	&0.8478	&0.0496	&0.2006 &  0.9636 \\ 
                    \midrule
                    {CORD19-OFT} & - & \textbf{0.2856} & \textbf{0.8606} & \textbf{0.0864} & \textbf{0.2417} & 0.9417 \\
                    
%-----------------------------------------------LEGAL------------------------------------------------------------
                    \hline
                    \multicolumn{7}{c}{\textit{{Law Domain}}} \\
                    \hline
                    \multirow{2}{*}{{Llama3-8B-Instruct}}
                    % &\makecell[l]{Adapter 1 \\ + Adapter 2} 
                    % & 0.2445 & 0.0768 & 0.1513 & 0.8488 & 0.0375 \\ 
                    &0 & 0.2360 & {0.8522} & 0.0210 & 0.1648 & \textbf{0.9255}\\ 
                    % &1 & 0.2188 & 0.8462 & 0.0148 & 0.1570 & 0.8899\\ 
                    &2 & 0.1985 & 0.7998 & 0.0114 & 0.1405 & 0.7514\\ 
                    \midrule
                    \multirow{1}{*}{{Llama3-70B-Instruct}}
                    &{0} & 0.2158	&0.8469	&0.0192	&0.1565 &  0.9123 \\ 
                    \midrule
                    {MultiLex-LoKr} & - & \textbf{0.2411} & \textbf{0.8526} & \textbf{0.0216} & \textbf{0.1679} & 0.9166 \\

%-----------------------------------------------NEWS-------------------------------------------------------------
                    \hline
                    \multicolumn{7}{c}{\textit{{News Domain}}} \\
                    \hline
                    \multirow{2}{*}{{Llama3-8B-Instruct}}
                    % &\makecell[l]{Adapter 1 \\ + Adapter 2} 
                    % & 0.2445 & 0.0768 & 0.1513 & 0.8488 & 0.0375 \\ 
                    &0 & 0.2488 & \textbf{0.8889} & 0.0912 & 0.4040 & 0.9768\\ 
                    % &1 & 0.2514 & 0.8849 & 0.0928 & 0.4096 & 0.9683\\ 
                    &{2} & {0.2569} & 0.8845 & {0.0953} & {0.4108} & 0.9726\\ 
                    \midrule
                    \multirow{1}{*}{{Llama3-70B-Instruct}}
                    &{0} & 0.2485	&0.8872	&0.0903	&0.4042 &  \textbf{0.9797} \\
                    \midrule
                    {Newsroom-LoKr} & - & \textbf{0.2976} & {0.8868} & \textbf{0.1283} & \textbf{0.4438} & 0.8909 \\
                    
                    \hline
                \end{tabular}
            }
            %\vspace{3mm}
            \caption{Baseline Performance of in-context learning from \texttt{Llama3-8B-Instruct} and \texttt{Llama3-70B-Instruct} compared with our trained Adapters on the Holdout validation dataset for all domains. k = number of examples used in In-context Learning.}
            \label{tab:icl_results}
        \end{table*}

\paragraph{Human Evaluation}

We recruited a total of eight domain experts—two for each domain from a local university and paid a fixed rate of 16 euros per hour. Annotators were enrolled in Master's programs in their respective fields, with the exception of the Medical domain, where licensed physicians participated in the study. A human evaluation was conducted on 100 × 4 summary samples, comprising 25 articles from each holdout validation dataset. Each article is paired with four summaries evaluated under two comparative settings: (i) Zero-shot vs. Cross-Domain, and (ii) Within-Domain vs. Cross-Domain. For each article, annotators were instructed to select one of the two summaries under both settings based on three criteria: correctness, conciseness, and coverage of the source content. In the Zero-shot vs. Cross-Domain comparison, annotators preferred cross-domain summaries in 2\%, 18\%, 16\%, and 10\% of cases for the Scientific, Medical, News, and Law domains, respectively. In the Within-Domain vs. Cross-Domain comparison, cross-domain summaries were rated as equally good or better than within-domain summaries in 12\%, 40\%, 10\%, and 64\% of cases across the same domains. We report the Top-1 agreement rate, defined as the proportion of instances in which both annotators selected the same summary as their top choice. The average Top-1 agreement across all datasets was 74.5\%, indicating substantial consistency between annotators.

\paragraph{Role of Dataset Similarity in Cross-Domain/With-Domain Inference}

An analysis of \autoref{tab:within_cross_scientific_unseen_test_data} and \autoref{tab:domain_similarity} reveals a strong correlation between dataset similarity and the downstream performance of the trained adapters. In each evaluated domain, Adapters trained on datasets exhibiting higher similarity with the holdout validation sets achieve superior results. For example, in the Scientific domain, the SciTLDR-AdaLoRA adapter achieves the best performance, which aligns with high dataset similarity between the SciTLDR training and scientific validation datasets. Likewise, in the Medical domain, adapters such as CORD19-OFT and MSLR-(IA)$^3$ demonstrate strong alignment with the Medical domain evaluation set. We see similar trends in the Legal and the news domain.  Collectively, these findings underscore the critical role of semantic and distributional similarity between training and inference datasets in maximizing model generalization and downstream task efficacy.

%In the Legal domain, the MultiLex-LoKr adapter, which is derived from a dataset with the highest contextual overlap, consistently outperforms alternatives, whereas adapters trained on EurLex—despite higher vocabulary overlap, are less effective, likely due to disparities in text length and structure. The News domain exhibits a similar trend, with Newsroom-LoKr achieving optimal performance due to its close distributional and contextual proximity to the validation data. In contrast, adapters trained in cross-domain settings on dissimilar datasets, exhibit reductions in evaluation performance.
\subsection{Comparison to Baselines}

As shown in \autoref{tab:icl_results}, our trained adapters outperform both \texttt{Llama3-8B-Instruct} and \texttt{Llama3-70B-Instruct} across most evaluation metrics and domains. They show consistent gains in the ROUGE, BLEU, and METEOR, and BERTScores, demonstrating superior relevance and quality in generated outputs compared to Llama3 baselines. However, it is noteworthy that the FActScore of our method, while strong, does not always surpass that of the baseline Llama3 models, particularly in some domains. This suggests that while our adapters enhance generation quality and domain adaptation, there may be trade-offs with factual consistency in certain cases. Overall, our method establishes clear improvements in the majority of metrics, especially in text quality and relevance, which highlights the effectiveness of domain specialization.

\section{Conclusion}

Through our experiments, we show that a smaller Large Language Model (LLM), such as \texttt{Llama-3-8B-Instruct}, can be effectively adapted to new domains when aided by PEFT methods. We introduce a comprehensive PEFT benchmark and find that while AdaLoRA demonstrates strong generalization across domains on the text summarization task, other PEFTs like LoKr and OFT may outperform it in specific domain contexts. Beyond single-domain adaptation, we investigate the effectiveness of combining multiple Adapters—each trained on different high-resource datasets to improve performance on previously unseen, low-resource domains. This approach reflects a practical and scalable strategy for real-world deployment, where domain-specific data may be scarce. Our analysis, conducted through inference on held-out validation sets under both Within-Domain and Cross-Domain settings, offers insights into strengths and limitations of both. Our results show that the Within-Domain setting consistently outperforms both the few-shot \texttt{Llama-3-8B-Instruct} and a much larger \texttt{Llama-3-70B-Instruct} baseline, underscoring the efficiency and adaptability of PEFTs. However, in the absence of in-domain high-resource datasets, we advocate for the use of Cross-Domain Adapters, which still yield competitive results and offer a practical fallback in low-resource conditions. We show that strategic combinations of PEFTs trained on multiple high-resource domains can enable robust generalization, even in highly constrained settings. This approach opens possibilities for reusable domain expertise and efficient deployment of LLMs across diverse applications.

%\section*{Acknowledgments}
\section*{Limitations}

Our experiments, while comprehensive, faced several constraints that should be acknowledged. We conducted our experiments on multiple datasets across various domains, evaluating the performance of Parameter-Efficient Fine-Tuning PEFT methods in both within-domain and cross-domain settings. However, due to time limitations, we focused our cross-domain evaluations on only the best-performing PEFTs. Our investigation encompassed six PEFT methods, notably excluding representation-efficient fine-tuning techniques, but not all possible PEFTs. Resource constraints restricted our training to five epochs and a maximum of one thousand samples per dataset. To accommodate context window limitations and resource constraints, we selected articles with the shortest lengths for training. Storage limitations confined our evaluation to a single Large Language Model (LLM). It is important to note that the effectiveness of PEFTs varied across datasets, and while we identified several practical approaches, we cannot definitively state which methods will underperform in specific settings, except in cases of poor-quality data. Interestingly, combining more than two PEFTs for inference often led to performance degradation. While our experiments span four domains and multiple PEFT methods, they are limited to a single LLM. Future work could explore generalization across a broader range of model architectures, tasks beyond summarization, and more linguistically and culturally diverse datasets. Additionally, further research is needed to understand how to best select, weight, or merge adapters for optimal performance across increasingly complex domain transfer scenarios.  
\section*{Ethical Statement}
Throughout our experiences, we have consistently adhered to the ACL Code of Ethics. Since we used established open-source benchmark datasets, privacy concerns were not relevant. The manual evaluation was carried out by our in-house domain specialists, who received an hourly wage of 16 euros. They were informed about the task and use of the data in the research, and their comments were stored anonymously to address any privacy concerns.

\noindent Our fine-tuning strategies ensured that no additional biases were introduced into the models beyond any existing biases in the model weights or in the reference dataset. However, we acknowledge that using high-resource domains as proxies for low-resource ones may inadvertently amplify biases present in the source domains, potentially skewing results or marginalizing target-domain perspectives.

\noindent Moreover, the chosen four domains may not comprehensively represent marginalized or linguistically diverse communities. Future work should investigate how adapter-based domain adaptation performs in truly underserved or low-resource linguistic contexts where domain boundaries may not align with high-resource analogs.

\noindent We also recognize a dual use risk inherent in adapter-based modular fine-tuning: the same techniques that enable efficient domain adaptation may be misused to generate convincing misinformation in sensitive areas such as legal or medical summarization. This risk is not addressed in our current evaluation but should be considered in broader deployments of such systems.

\noindent The research aims to evaluate if using Adapters can improve LLM performance on unseen domains. The results and discussions presented in this article aim to advance research on domain-specific text summarization with a focus on low-resource domains. All training scripts and trained models are made available to the research community.
\bibliography{custom}

\appendix

\section{Prompts}
\label{subsec:prompts}
        \begin{itemize}
            \item Scientific Domain: \textit{Summarize the provided scientific article in a clear and concise paragraph. Include the study’s objective, background, methodology, key findings, and conclusions, ensuring the summary represents the article’s essence.}
            \item Medical Domain: \textit{Provide a cohesive summary of the given medical article in one paragraph. Highlight the study’s objective, background, methods, major conclusions, and potential clinical implications in a clear and professional manner.}
            \item Legal Domain: \textit{Generate a concise paragraph summarizing the provided legal case study. Address the background, legal questions, key arguments, rulings, and any significant precedents, ensuring clarity and accuracy.}
            \item News Domain: \textit{Create a clear and concise summary of the given news article in one paragraph. Focus on the main event, its context, key details, and broader implications, presenting the information in an informative manner.}
        \end{itemize}
        \section{Technical Details}
        We used A100 80GB GPUs for the training and evaluation of Adapters.
        \label{app:tech-details}
    \subsection{Training Hyperparameters}
    \label{subsec:train_hyp_para}
    We trained each PEFT attached to Llama3 for five epochs and stored the best-performing checkpoint. We used a context window of 4096 and set returning\_overflowing\_tokens=True for datasets with contexts longer than 4096, allowing the model to see the entire content in one batch as we set the batch size to 4. We maintained constant model-specific training hyperparameters across all fine-tunings. We set the learning rate to 0.0005, with cosine learning rate decay at bfloat16 precision. We trained and validated all combinations on 1000 training samples and 500 validation samples. We altered the hyperparameters for the PEFTs we used, and we list them below:
    \begin{enumerate}
        \item \textbf{AdaLoRA}: AdaLoRA uses a low-rank adaptation mechanism, so we set the low rank to 64 with alpha=8 and lora\_dropout=0.01.
        \item \textbf{(IA)$^{3}$}: (IA)$^{3}$ had no specific hyperparameters except that they were initialized with random weights.
        \item \textbf{LoHA}: Another low-rank method with the rank set to 64, alpha=8, rank\_dropout=0.1, and module\_dropout=0.
        \item \textbf{LoKr}: The hyperparameters were the same as LoHA, with the low rank set to 64, alpha=8, rank\_dropout=0.1, and module\_dropout=0.
        \item \textbf{LoRA}: The first peft with which we experimented and found out that rank 64 gave better results than any other rank, with lora\_alpha=8, lora\_dropout=0.1 and only the bias of the LoRA module was allowed to update\footnote{\href{https://shorturl.at/xugiX}{huggingface.co/docs/peft/lora/bias}}. 
        \item \textbf{OFT}: The inner matrices of OFT were set to have a rank of 64, and with a module\_dropout of 0.0, were the only necessary hyperparameters of this PEFT. 
    \end{enumerate}

\section{Performance of Parameter-Efficient-Fine-tuning-Techniques}
\label{app:peft-performance}
\begin{figure*}[!htbp]
  \includegraphics[width=\textwidth]{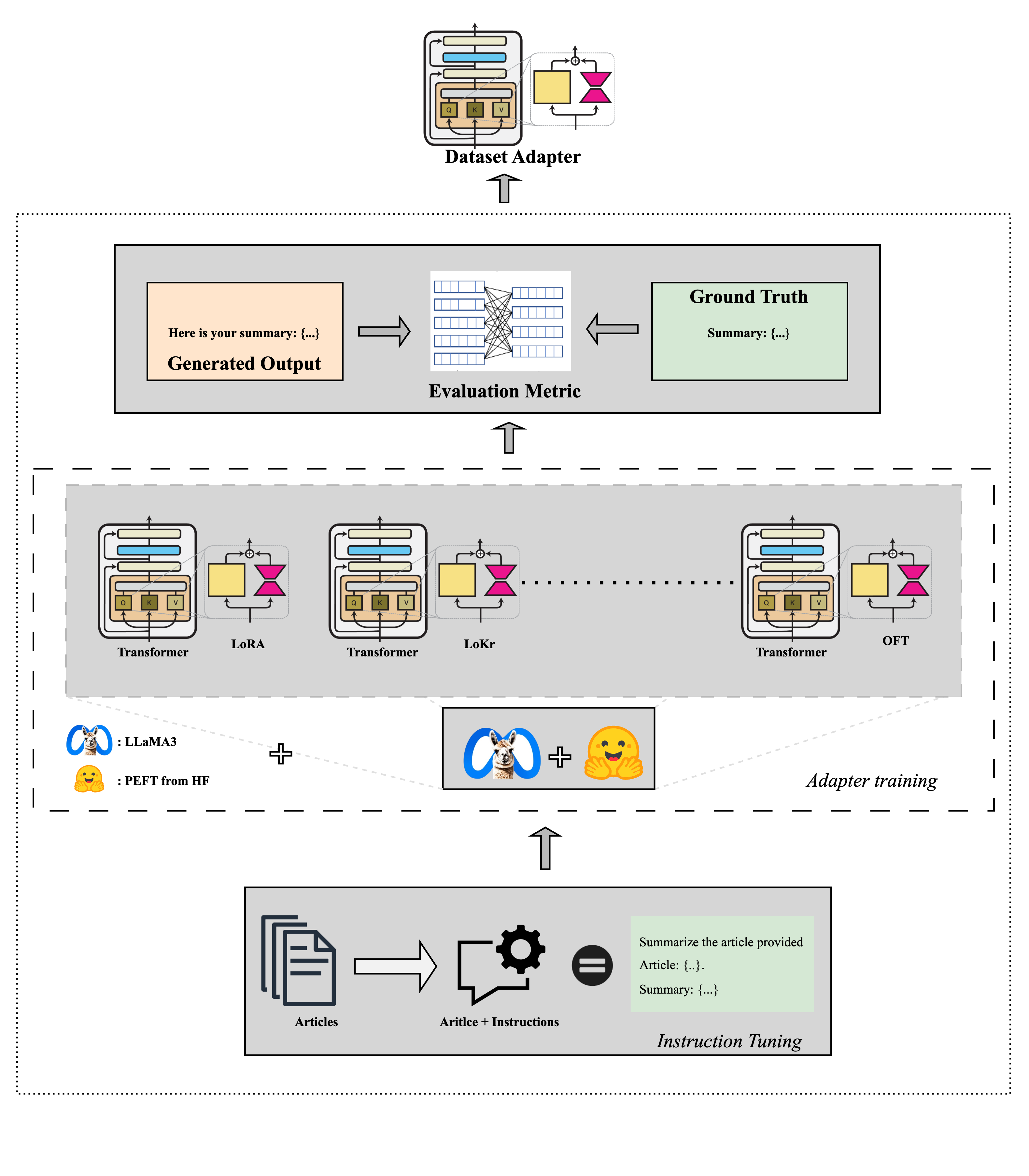}
  \caption{Training and Evaluation pipeline for PEFT selection for each dataset. A similar approach is used for dataset selection to ensure it is domain representative. Select Domain Adapters are later used for experiments on our Holdout Validation datasets under the Within-Domain and Cross-Domain settings. }
  \label{fig:peft-selection}
\end{figure*}

\begin{figure*}[!htbp]
  \includegraphics[width=\textwidth]{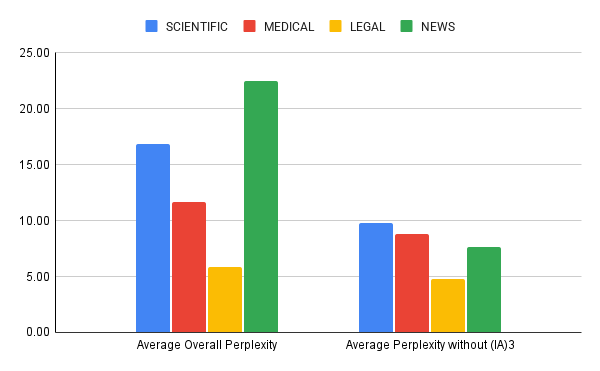}
  \caption{Average Perplexity Scores per domain computed during fine-tuning. }
  \label{fig:perlexity}
\end{figure*}

We illustrate the pipeline used for evaluating PEFT in \autoref{fig:peft-selection}. During fine-tuning, we evaluate them using the Perplexity scores and post-training, we use the performance on the test datasets of the respective datasets to rank them. The perplexity scores are outlined in \autoref{fig:perlexity}. In general, a lower perplexity indicates better training. Surprisingly, we recorded a higher perplexity score with $(IA)^3$, indicating a bad training, however, we noticed that post-training, Adapter trained using $(IA)^3$ were given more preference using automatic and human evaluation. This indicates that perplexity as a metric standalone is not reliable and should not be used as the sole indicator of performance.

\paragraph{Evaluation of PEFTs}
\label{app:peft:ranks}

The individual ranks of all PEFTs on all datasets are shown in \autoref{tab:pefts_rank_full}.
\begin{table*}[htbp]
    \centering
    \small
    % \scalebox{0.92}
    \resizebox{\textwidth}{!}
    {    
        \setlength{\tabcolsep}{3pt} % Adjust column spacing
            \begin{tabular}{llcccccc}
                \hline
                \textbf{Domain} & \textbf{Dataset} & \textbf{AdaLora} & \textbf{(IA)$^3$} & \textbf{LoHA} & \textbf{LoKr} & \textbf{LoRA} & \textbf{OFT} \\ \hline
                
                \multicolumn{2}{c}{\multirow{2}{*}{\textbf{Trainable Parameters}}} &54,534,144 & 786,432 & 109,051,904 & 786,432 & 54,525,952 & 17,825,792 \\ 
                 \multicolumn{2}{c}{} & (0.6745\%) & (0.0098\%) & (1.3398\%) & (0.0098\%) & (0.6744\%) & (0.2215\%) \\
                \hline
                % SCIENTIFIC ROWS
                \multirow{3}{*}{Scientific} 
                & Arxiv         & {4} & {6}  & {2} & \textbf{1}  & {3} & {5} \\ %\cline{2-8} 
                & Elsevier      & \textbf{1} & {2}  & {4} & {3}  & {6} & {5} \\ %\cline{2-8} 
                & SciTLDR       & \textbf{1} & {2}  & {4} & {3}  & {6} & {5} \\ \hline 
                
                % MEDICAL ROWS
                \multirow{4}{*}{Medical} 
                & CORD19        & \textbf{1} & {5}  & {6} & {4}  & {3} & {2} \\ %\cline{2-8}
                & MSLR          & {2} & \textbf{1}  & {3} & {5}  & {4} & {6} \\ %\cline{2-8}
                & PubMed        & {5} & {6}  & {4} & {3}  & \textbf{1} & {2} \\ %\cline{2-8}
                & Sci Lay       & {4} & {5}  & {6} & {2}  & {3} & \textbf{1} \\ \hline
                
                % LEGAL ROWS
                \multirow{3}{*}{Legal} 
                & BillSum       & {5} & {6}  & {4} & \textbf{1}  & {3} & {2} \\ %\cline{2-8}
                & Eur-Lex-Sum   & {6} & {4}  & {2} & {5}  & {3} & \textbf{1} \\ %\cline{2-8} 
                & Multi-Lex     & {6} & {5}  & {2} & {4}  & \textbf{1} & {3} \\ \hline

                % NEWS ROWS
                \multirow{4}{*}{News} 
                & CNN/DM            & \textbf{1} & {3}  & {4} & {2}  & {6} & {5} \\ %\cline{2-8} 
                & Multi News        & {3} & {6}  & {4} & {5}  & {2} & \textbf{1} \\ %\cline{2-8} 
                & Newsroom          & \textbf{1} & {4}  & {5} & {2}  & {3} & {6} \\ %\cline{2-8}
                & XSum              & {2} & {4}  & {3} & \textbf{1}  & {6} & {5}\\ \hline 
                
            \end{tabular}
        }
    %\vspace{3mm}
    \caption{Comparison of Trainable Parameters across Domains for all PEFTs and their respective ranks according to Borda Count Method such that 1 = best performance and 6 = worst performance. See Appendix \ref{app:peft:ranks} for all ranks and detailed scores.}
    \label{tab:pefts_rank_full}
\end{table*}

\paragraph{Scientific Domain}
    We summarize the performance of the three datasets scientific domain in \autoref{tab:scientific_test_split}.
        \begin{table*}[ht]
            \centering
            % \scalebox{1}
            \resizebox{\linewidth}{!}
            {
                \begin{tabular}{llrrrrrr}
                    \hline
                     \textbf{Test Dataset} & \textbf{PEFT} & \multicolumn{3}{c}{\textbf{ROUGE}} & \textbf{BERTScore} & \textbf{BLEU} & \textbf{METEOR}\\
                            & & \multicolumn{1}{c}{\textbf{Rouge1}} & \multicolumn{1}{c}{\textbf{Rouge2}} & \multicolumn{1}{c}{\textbf{RougeL}} & & & \\
                     \hline
                     \multirow{7}{*}{Arxiv} 
                    & Zero-Shot     & \textit{0.3799} & \textit{0.1404} & \textit{0.2184} & \textit{0.8400} & \textit{0.0608} & \textit{0.3057} \\
                    & AdaLoRA       & 0.3889 & 0.1493 & 0.2250 & 0.8506 & 0.1032 & 0.3348 \\
                    & (IA)$^3$      & 0.3854 & 0.1430 & 0.2236 & 0.8408 & 0.0636 & 0.3148 \\
                    & LoHA          & \textbf{0.3900} & 0.1498 & 0.2256 & \textbf{0.8516} & 0.1054 & 0.3404 \\
                    & \textbf{LoKr} & 0.3893 & \textbf{0.1520} & \textbf{0.2313} & 0.8514 & \textbf{0.1105} & \textbf{0.3449} \\
                    & LoRA          & 0.3877 & 0.1518 & 0.2292 & 0.8466 & 0.1087 & 0.3404 \\
                    & OFT           & 0.3801 & 0.1487 & 0.2290 & 0.8463 & 0.1053 & 0.3332  \\
                    \hline
        
                    \multirow{7}{*}{Elsevier} 
                    & Zero-Shot     & \textit{0.4027} & \textit{0.1391} &\textit{ 0.2252} & \textit{0.8645} & \textit{0.0993} & \textbf{\textit{0.3215}} \\
                    & \textbf{AdaLoRA} & \textbf{0.4178} & \textbf{0.1580} & \textbf{0.2401} & 0.8637 & \textbf{0.1164} & 0.3211  \\
                    & (IA)$^3$      & 0.4063 & 0.1411 & 0.2267 & \textbf{0.8650} & 0.0994 & 0.3111  \\
                    & LoHA          & 0.3897 & 0.1463 & 0.2244 & 0.8542 & 0.1049 & 0.3129 \\
                    & LoKr          & 0.3874 & 0.1448 & 0.2303 & 0.8548 & 0.1083 & 0.3062 \\
                    & LoRA          & 0.3816 & 0.1387 & 0.2215 & 0.8495 & 0.1048 & 0.3091 \\
                    & OFT           & 0.3841 & 0.1396 & 0.2187 & 0.8510 & 0.1032 & 0.3106  \\
                    \hline
        
                    \multirow{7}{*}{SciTLDR} 
                    & Zero-Shot     & \textit{0.3425} & \textit{0.1009} & \textit{0.1961} & \textit{0.8677} & \textit{0.0511} & \textit{0.2731} \\
                    & \textbf{AdaLoRA} & 0.3436 & \textbf{0.1125} & \textbf{0.2168} & \textbf{0.8692} & \textbf{0.0794} & 0.2571 \\
                    & (IA)$^3$      & \textbf{0.3448} & 0.1007 & 0.1958 & 0.8676 & 0.0503 & \textbf{0.3189} \\
                    & LoHA          & 0.2836 & 0.0844 & 0.1805 & 0.8513 & 0.0400 & 0.2732 \\
                    & LoKr          & 0.3254 & 0.1035 & 0.2066 & 0.8658 & 0.0739 & 0.2546 \\
                    & LoRA          & 0.2270 & 0.0659 & 0.1518 & 0.8371 & 0.0319 & 0.2748 \\
                    & OFT           & 0.2281 & 0.0661 & 0.1553 & 0.8361 & 0.0320 & 0.2754 \\
                    \hline
                \end{tabular}
            }
            %\vspace{3mm}
            \caption{Performance of PEFTs trained on scientific domain datasets and inferred on the respective test data split.}
            \label{tab:scientific_test_split}
        \end{table*}

\paragraph{Medical Domain}
 The performance of individual and combinations PEFTs fine-tuned on the medical domain datasets is provided in the \autoref{tab:medical_test_split}.

\renewcommand{\arraystretch}{1.1}
        \begin{table*}[!h]
            \centering
            % \scalebox{1}
            \resizebox{\linewidth}{!}
            {
                \begin{tabular}{llrrrrrr}
                    \hline
                     \textbf{Test Dataset} & \textbf{PEFT} & \multicolumn{3}{c}{\textbf{ROUGE}} & \textbf{BERTScore} & \textbf{BLEU} & \textbf{METEOR}\\
                            & & \multicolumn{1}{c}{\textbf{Rouge1}} & \multicolumn{1}{c}{\textbf{Rouge2}} & \multicolumn{1}{c}{\textbf{RougeL}} & & & \\
                     \hline
                     \multirow{7}{*}{CORD-19} 
                    & Zero-Shot     & \textit{0.4159} & \textit{0.1618} & \textit{0.2315} & \textit{0.8625} & \textit{0.1220} & \textit{0.2971} \\
                    & \textbf{AdaLoRA}       & \textbf{0.4315} & 0.2006 & \textbf{0.2681} & \textbf{0.8633} & 0.1794 & \textbf{0.3212} \\
                    & (IA)$^3$      & 0.4156 & 0.1631 & 0.2319 & 0.8632 & 0.1223 & 0.2961 \\
                    & LoHA          & 0.4094 & 0.1834 & 0.2463 & 0.8565 & 0.1668 & 0.3114 \\
                    & LoKr          & 0.4169 & 0.1934 & 0.2595 & 0.8599 & 0.1763 & 0.3131 \\
                    & LoRA          & 0.4176 & 0.1987 & 0.2607 & 0.8582 & 0.1831 & 0.3175 \\
                    & OFT           & 0.4175 & \textbf{0.2013} & 0.2634 & 0.8580 & \textbf{0.1901} & 0.3152 \\
                    \hline
        
                    \multirow{7}{*}{MSLR} 
                    & Zero-Shot     & \textit{0.1644} & \textit{0.0298} & \textit{0.1110} & \textit{0.8278} & \textit{0.0121} & \textit{0.1958} \\
                    & AdaLoRA       & 0.1355 & 0.0237 & 0.1016 & 0.8189 & 0.0095 & 0.1767 \\
                    & \textbf{(IA)$^3$}      & \textbf{0.1674} & \textbf{0.0310} & \textbf{0.1132} & \textbf{0.8297} & \textbf{0.0122} & \textbf{0.1998} \\
                    & LoHA          & 0.1093 & 0.0183 & 0.0918 & 0.8075 & 0.0080 & 0.1525 \\
                    & LoKr          & 0.1066 & 0.0161 & 0.0866 & 0.8052 & 0.0063 & 0.1406 \\
                    & LoRA          & 0.1065 & 0.0171 & 0.0899 & 0.8074 & 0.0071 & 0.1504 \\
                    & OFT           & 0.1027 & 0.0153 & 0.0865 & 0.8035 & 0.0066 & 0.1415 \\
                    \hline
        
                    \multirow{7}{*}{PubMed} 
                    & Zero-Shot     & \textit{0.3780} & \textit{0.1495} & \textit{0.2228} & \textit{0.8460} & \textit{0.0855} & \textit{0.3269} \\
                    & AdaLoRA       & 0.3874 & 0.1733 & 0.2422 & 0.8532 & 0.1419 & 0.3525 \\
                    & (IA)$^3$      & 0.3845 & 0.1528 & 0.2272 & 0.8468 & 0.0868 & 0.3315 \\
                    & LoHA          & 0.3852 & 0.1769 & 0.2441 & 0.8557 & 0.1610 & 0.3572 \\
                    & LoKr          & 0.3824 & 0.1782 & 0.2466 & 0.8544 & 0.1579 & 0.3591 \\
                    & \textbf{LoRA}          & \textbf{0.3954} & \textbf{0.1916} & \textbf{0.2604} & 0.8589 & \textbf{0.1715} & 0.3727 \\
                    & OFT           & 0.3919 & 0.1877 & 0.2564 & \textbf{0.8596} & 0.1665 &\textbf{0.3757} \\
                    \hline
        
                    \multirow{7}{*}{SciLay} 
                    & Zero-Shot     & \textit{0.3627} & \textit{0.1224} & \textit{0.2041} & \textit{0.8593} & \textit{0.0776} & \textit{0.3112} \\
                    & AdaLoRA       & 0.3658 & 0.1333 & 0.2089 & 0.8584 & 0.0934 & 0.3202 \\
                    & (IA)$^3$      & 0.3645 & 0.1231 & 0.2038 & \textbf{0.8596} & 0.0777 & 0.3109 \\
                    & LoHA          & 0.3555 & 0.1254 & 0.2005 & 0.8534 & 0.0841 & 0.3164 \\
                    & LoKr          & \textbf{0.3690} & 0.1347 & \textbf{0.2148} & 0.8586 & 0.0960 & 0.3163 \\
                    & LoRA          & 0.3642 & 0.1335 & 0.2119 & 0.8567 & 0.0954 & \textbf{0.3360} \\
                    & \textbf{OFT}           & 0.3631 & \textbf{0.1362} & 0.2120 & 0.8568 & \textbf{0.0982} & 0.3270 \\
                    \hline
                \end{tabular}
            }
            %\vspace{3mm}
            \caption{Performance of PEFTs trained on medical domain datasets and inferred on the respective test data split.}
            \label{tab:medical_test_split}
        \end{table*}

\paragraph{Legal Domain}
 The performance of individual and combinations PEFTs fine-tuned is illustrated in \autoref{tab:legal_test_split}.
        \renewcommand{\arraystretch}{1.1}
        \begin{table*}[ht]
            \centering
            % \scalebox{1}
            \resizebox{\linewidth}{!}
            {
                \begin{tabular}{llrrrrrr}
                    \hline
                     \textbf{Test Dataset} & \textbf{PEFT} & \multicolumn{3}{c}{\textbf{ROUGE}} & \textbf{BERTScore} & \textbf{BLEU} & \textbf{METEOR}\\
                            & & \multicolumn{1}{c}{\textbf{Rouge1}} & \multicolumn{1}{c}{\textbf{Rouge2}} & \multicolumn{1}{c}{\textbf{RougeL}} & & & \\
                     \hline
                     \multirow{7}{*}{BillSum} 
                    & Zero-Shot     & \textit{0.4491} & \textit{0.2201} & \textit{0.2947} & \textit{0.8697} & \textit{0.1582} & \textit{0.2936} \\
                    & AdaLoRA       & 0.5164 & 0.3109 & 0.3784 & \textbf{0.8856} & 0.2540 & 0.4321 \\
                    & (IA)$^3$      & 0.4453 & 0.2128 & 0.2900 & 0.8676 & 0.1552 & 0.3779 \\
                    & LoHA          & 0.5093 & 0.3262 & 0.3791 & 0.8833 & 0.2608 & 0.4719 \\
                    & \textbf{LoKr} & \textbf{0.5178} & \textbf{0.3351} & \textbf{0.3903} & 0.8853 & \textbf{0.2702} & \textbf{0.4744} \\
                    & LoRA          & 0.5034 & 0.3265 & 0.3823 & 0.8837 & 0.2634 & 0.4694 \\
                    & OFT           & 0.5074 & 0.3296 & 0.3850 & 0.8837 & 0.2647 & 0.4714 \\
                    \hline
        
                    \multirow{7}{*}{Eur-Lex} 
                    & Zero-Shot     & \textit{0.2848} & \textit{0.1327} & \textit{0.1640} & \textit{0.8466} & \textit{0.0071} & \textit{0.1239} \\
                    & AdaLoRA       & 0.2024 & 0.0661 & 0.1481 & 0.7638 & 0.0019 & 0.0767 \\
                    & (IA)$^3$      & 0.2852 & 0.1322 & 0.1640 & 0.8462 & 0.0071 & \textbf{0.1247} \\
                    & LoHA          & 0.2898 & 0.1623 & 0.1837 & 0.8674 & 0.0075 & 0.1240 \\
                    & LoKr          & 0.2859 & 0.1619 & 0.1811 & 0.8613 & 0.0071 & 0.1231 \\
                    & LoRA          & 0.2901 & 0.1598 & 0.1848 & 0.8652 & 0.0074 & 0.1228 \\
                    & \textbf{OFT}           & \textbf{0.2936} & \textbf{0.1658} & \textbf{0.1859} & \textbf{0.8700} & \textbf{0.0076} & 0.1245 \\
                    \hline
        
                    \multirow{7}{*}{Multi-Lex} 
                    & Zero-Shot     & \textit{0.3883} & \textit{0.1571} & \textit{0.2264} & \textit{0.8499} & \textit{0.1234} & \textit{0.2330} \\
                    & AdaLoRA       & 0.3435 & 0.1496 & 0.2155 & 0.8334 & 0.1196 & 0.2621 \\
                    & (IA)$^3$      & 0.3940 & 0.1570 & 0.2280 & 0.8511 & 0.1239 & 0.2986 \\
                    & LoHA          & 0.4539 & 0.2403 & 0.3067 & 0.8618 & 0.2075 & 0.3583 \\
                    & LoKr          & 0.4367 & 0.2205 & 0.2899 & 0.8564 & 0.1859 & 0.3354 \\
                    & \textbf{LoRA}          & \textbf{0.4578} & \textbf{0.2474} & \textbf{0.3137} & \textbf{0.8659} & \textbf{0.2157} & \textbf{0.3586} \\
                    & OFT           & 0.4471 & 0.2443 & 0.3109 & 0.8545 & 0.2120 & 0.3529 \\
                    \hline
                \end{tabular}
            }
            %\vspace{3mm}
            \caption{Performance of PEFTs trained on legal domain datasets and inferred on the respective test data split.}
            \label{tab:legal_test_split}
        \end{table*}

\paragraph{News Domain}
The inference scores on the four datasets from the News domain are outline in \autoref{tab:news_test_split}.
        
        \renewcommand{\arraystretch}{1.1}
        \begin{table*}[ht!]
            \centering
            % \scalebox{1}
            \resizebox{\linewidth}{!}
            {
                \begin{tabular}{llrrrrrr}
                    \hline
                     \textbf{Test Dataset} & \textbf{PEFT} & \multicolumn{3}{c}{\textbf{ROUGE}} & \textbf{BERTScore} & \textbf{BLEU} & \textbf{METEOR}\\
                            & & \multicolumn{1}{c}{\textbf{Rouge1}} & \multicolumn{1}{c}{\textbf{Rouge2}} & \multicolumn{1}{c}{\textbf{RougeL}} & & & \\
                     \hline
                     \multirow{7}{*}{CNN/DM} 
                    & Zero-Shot     & \textit{0.3311} & \textit{0.1390} & \textit{0.2144} & \textit{0.8727} & \textit{0.0638} & \textit{0.3930} \\
                    & \textbf{AdaLoRA}       & \textbf{0.3912} & \textbf{0.1783} & \textbf{0.2562} & \textbf{0.8785} & \textbf{0.0951} & \textbf{0.4384} \\
                    & (IA)$^3$      & 0.3338 & 0.1375 & 0.2120 & 0.8732 & 0.0623 & 0.3981 \\
                    & LoHA          & 0.2533 & 0.1307 & 0.1821 & 0.8559 & 0.0675 & 0.3951 \\
                    & LoKr          & 0.3019 & 0.1498 & 0.2153 & 0.8632 & 0.0775 & 0.4032 \\
                    & LoRA          & 0.2384 & 0.1198 & 0.1752 & 0.8519 & 0.0637 & 0.3697 \\
                    & OFT           & 0.2541 & 0.1309 & 0.1819 & 0.8555 & 0.0701 & 0.3940 \\
                    \hline
        
                    \multirow{7}{*}{Multinews} 
                    & Zero-Shot     & \textit{0.4048} & \textit{0.1317} & \textit{0.2110} & \textit{0.8661} & \textit{0.0743} & \textit{0.2657} \\
                    & AdaLoRA       & 0.4169 & 0.1489 & 0.2248 & \textbf{0.8673} & 0.0837 & 0.2748 \\
                    & (IA)$^3$      & 0.4001 & 0.1317 & 0.2132 & 0.8659 & 0.0736 & 0.2641 \\
                    & LoHA          & 0.4398 & 0.1726 & 0.2343 & 0.8660 & 0.1365 & 0.3522 \\
                    & LoKr          & 0.4368 & 0.1721 & 0.2386 & 0.8642 & 0.1372 & 0.3507 \\
                    & LoRA          & \textbf{0.4451} & 0.1828 & 0.2452 & 0.8662 & 0.1512 & \textbf{0.3609} \\
                    & \textbf{OFT}           & 0.4429 & \textbf{0.1834} & \textbf{0.2460} & 0.8666 & \textbf{0.1519} & 0.3579 \\
                    \hline
        
                    \multirow{7}{*}{Newsroom} 
                    & Zero-Shot     & \textit{0.1936} & \textit{0.0759} & \textit{0.1380} & \textit{0.8589} & \textit{0.0328} & \textit{0.2370} \\
                    & \textbf{AdaLoRA}       & \textbf{0.2609} & 0.1356 & \textbf{0.2071} & \textbf{0.8649} & \textbf{0.0790} & \textbf{0.3346} \\
                    & (IA)$^3$      & 0.1899 & 0.0731 & 0.1364 & 0.8589 & 0.0326 & 0.2702 \\
                    & LoHA          & 0.1615 & 0.0979 & 0.1341 & 0.8382 & 0.0457 & 0.2909 \\
                    & LoKr          & 0.2329 & \textbf{0.1429} & 0.1913 & 0.8584 & 0.0739 & 0.3446 \\
                    & LoRA          & 0.1551 & 0.1044 & 0.1357 & 0.8322 & 0.0511 & 0.3024 \\
                    & OFT           & 0.1505 & 0.1013 & 0.1319 & 0.8306 & 0.0491 & 0.2915 \\
                    \hline
        
                    \multirow{7}{*}{XSum} 
                    & Zero-Shot     & \textit{0.1660} & \textit{0.0390} & \textit{0.1134} & \textit{0.8615} & \textit{0.0115} & \textit{0.3067} \\
                    & AdaLoRA       & 0.2267 & 0.0774 & 0.1648 & \textbf{0.8743} & 0.0288 & 0.2983 \\
                    & (IA)$^3$      & 0.1651 & 0.0396 & 0.1138 & 0.8613 & 0.0124 & 0.2367 \\
                    & LoHA          & 0.2133 & 0.0762 & 0.1562 & 0.8684 & 0.0258 & 0.2956 \\
                    & \textbf{LoKr} & \textbf{0.2380} & \textbf{0.0873} & \textbf{0.1756} & 0.8730 & \textbf{0.0351} & \textbf{0.3097} \\
                    & LoRA          & 0.1094 & 0.0388 & 0.0813 & 0.8375 & 0.0123 & 0.2059 \\
                    & OFT           & 0.1303 & 0.0480 & 0.0986 & 0.8454 & 0.0137 & 0.2252 \\
                    \hline
                \end{tabular}
            }
            %\vspace{2mm}
            \caption{Performance of PEFTs trained on news domain datasets and inferred on the respective test data split.}
            \label{tab:news_test_split}
        \end{table*}

\paragraph{Domain Adapters}
\label{app:domain-adapters}

Following the methodology outlined in \autoref{fig:methodology}, we evaluate the performance of all individual and combination dataset adapters under Within-Domain settings. The top-3 performing adapters for each domain are presented in \autoref{tab:news_test_split_within_across}. These results are based on rankings derived using the Borda Count Method, which aggregates performance across multiple metrics.

\noindent Our findings reveal that, across all domains, the top-3 ranked adapters are individual (single) adapters rather than combinations. This suggests that, contrary to expectations, merging multiple dataset-specific adapters within the same domain does not lead to significant performance improvements. In fact, individual adapters often capture domain-specific patterns more effectively, likely due to reduced interference and greater specialization. These observations indicate that, for Within-Domain applications, a well-chosen single adapter may suffice.

\begin{table*}[h]
    \centering
    % \scalebox{1}
    %\resizebox{\linewidth}{!}
    {
    \begin{tabular}{llcccc}
    \hline
    \textbf{Domain} & \textbf{Adapters} & \multicolumn{1}{c}{\textbf{ROUGE}} & \textbf{BERTScore} & \textbf{BLEU} & \textbf{METEOR}\\

        \toprule
        \multirow{3}{*}{Scientific} 
        
        & \textbf{Arxiv-LoKr} & 0.2392 & 0.8514 & 0.1105 & \textbf{0.3449} \\ %\cline{2-6}
        % & \makecell[l]{Adapter1 \\ + Adapter2}  & 0. & 0. & &  \\ \cline{2-6}
        & \makecell[l]{Elsevier-AdaLoRA}  & \textbf{0.2512} & 0.8637 & \textbf{0.1164} & 0.3211 \\ %\cline{2-6}
        % & \makecell[l]{Adapter1 \\ + Adapter2}  &  & &  & \\ \toprule
        & \makecell[l]{SciTLDR-AdaLoRA}  & 0.2031 & \textbf{0.8692} & 0.0794 & 0.3189 \\ \toprule
        
        \multirow{4}{*}{Medical} 
        % & \makecell[l]{Adapter1 \\ + Adapter2} & 0. & 0. & 0. & 0. \\ \cline{2-6}
        % & \makecell[l]{Adapter1 \\ + Adapter2}  & 0. & 0. & 0. & 0. \\ \cline{2-6}
        % & \makecell[l]{Adapter1 \\ + Adapter2}  & 0. & 0. & 0. & 0. \\ 
        & \makecell[l]{CORD19-AdaLoRA} & \textbf{0.2852} & \textbf{0.8633} & \textbf{0.1794} & 0.3212 \\ %\cline{2-6}
        % & \makecell[l]{MSLR-(IA)$^3$}  & 0.0838 & 0.8297 & 0.0122 & 0.1998 \\ \cline{2-6}
        & \makecell[l]{\textbf{PubMed-LoRA}}  & 0.2702 & 0.8589 & 0.1715 & \textbf{0.3727} \\ %\cline{2-6}
        & \makecell[l]{SciLay-OFT}  & 0.2202 & 0.8596 & 0.0982 & 0.3270 \\
        \toprule
        
        \multirow{3}{*}{Law} 
        % & \textbf{Adapter1} & \textbf{0.2817} & \textbf{0.1555} &  \textbf{0.0930} & \textbf{0.3588} \\ \cline{2-6}
        % & \makecell[l]{Adapter1 \\ + Adapter2} & 0.1033 & 0.0446 & 0.0822 & 0.8221  \\ \cline{2-6}
        % & \makecell[l]{Adapter1\\ + Adapter2} & 0.1342 & 0.0836 & 0.1158  &0.8320 \\ 
        & \textbf{BillSum-LoKr} & \textbf{0.4076} & \textbf{0.8853} &  \textbf{0.2702} & \textbf{0.4744} \\ %\cline{2-6}
        & \makecell[l]{EurLex-OFT} & 0.2084 & 0.8700 & 0.0076 & 0.1245  \\ %\cline{2-6}
        & \makecell[l]{MultiLex-LoRA} & 0.3287 & 0.8659 & 0.2157  &0.3586 \\ 
        \toprule

        \multirow{4}{*}{News} 
        % & \textbf{Adapter1} & \textbf{0.2451} & \textbf{0.0880}  & \textbf{0.1834} & \textbf{0.8748} \\ \cline{2-6}
        % & \makecell[l]{Adapter1 \\ + Adapter2} & 0.0888 & 0.0263 & 0.0715 & 0.8323  \\ \cline{2-6}
        % & \makecell[l]{Adapter1 \\ + Adapter2 \\ + Adapter3} & 0.1040 & 0.0355 & 0.0829 & 0.8288 \\
        & \makecell[l]{\textbf{CNNDM-AdaLoRA}} & 0.2614 & \textbf{0.8785} & 0.0951 & \textbf{0.4384} \\ %\cline{2-6}
        & \makecell[l]{Multinews-OFT}  & \textbf{0.2714} & 0.8666 & \textbf{0.1519} & 0.3579 \\% \cline{2-6}
        & \makecell[l]{Newsroom-AdaLoRA}  & 0.1942 & 0.8649 & 0.0790 & 0.3346 \\ %\cline{2-6}
        % & \makecell[l]{XSum-LoKr}  & 0.1540 & 0.8730 & 0.0351 & 0.3097 \\
         \toprule           
                    
        \end{tabular}
        }
            %\vspace{3mm}
        \caption{Table showing the top-3 \textit{Dataset-Adapters} among individual and combination Adapters for each domain dataset. The bold values are the highest scores (and they surpassed zero-shot scores on the respective datasets).} % \textcolor{red}{@mehul: include numbers / best-performing adapters here from the test sets of the dataset}}
        \label{tab:news_test_split_within_across}
        \end{table*}

\end{document}